%% file: main.tex
\documentclass[conference]{IEEEtran}
\IEEEoverridecommandlockouts

\usepackage{amsmath,amssymb,amsfonts}
\usepackage{algorithmic}
\usepackage{graphicx, subcaption}
\usepackage{textcomp}
\usepackage{xcolor}

\usepackage{minted}
\usepackage{listings}

\lstdefinelanguage{json}{
  basicstyle=\ttfamily\footnotesize,
  numbers=left,
  numberstyle=\tiny,
  stepnumber=1,
  numbersep=8pt,
  showstringspaces=false,
  breaklines=true,
  frame=single,
  backgroundcolor=\color{gray!10},
  string=[s]{"}{"},
  comment=[l]{:},
  morestring=[b]',
}

\usepackage{pythontex}

\usepackage[most]{tcolorbox}

\tcbset{textmarker/.style={%
        enhanced,
        parbox=false,boxrule=0mm,boxsep=0mm,arc=0mm,
        outer arc=0mm,left=6mm,right=3mm,top=7pt,bottom=7pt,
        toptitle=1mm,bottomtitle=1mm,oversize}}

\newtcolorbox{hintBox}{textmarker,
    borderline west={6pt}{0pt}{yellow},
    colback=yellow!10!white}
\newtcolorbox{importantBox}{textmarker,
    borderline west={6pt}{0pt}{red},
    colback=red!10!white}
\newtcolorbox{noteBox}{textmarker,
    borderline west={6pt}{0pt}{green},
    colback=green!10!white}

\newtcolorbox{taskBox}{textmarker,
    colback=blue!10!white,
    left=5pt, right=5pt, top=5pt,bottom=5pt}

\newcommand{\task}[1]{\begin{taskBox} \textbf{Task:} #1 \end{taskBox}}

\definecolor{iccvblue}{rgb}{0.21,0.49,0.74}
\usepackage[pagebackref,breaklinks,colorlinks,allcolors=iccvblue]{hyperref}

\usepackage{cite}

\makeatletter
\def\citepunct{,\penalty\@m\ } 
\makeatother
\renewcommand\IEEEkeywordsname{Keywords}

\title{\LARGE Driving scenario generation and evaluation using a structured layer representation and foundational models\\
\thanks{We thank Google Cloud for funding this project by providing access to computational credits for Vertex AI and Gemini.}
}

\author{Arthur Hubert, Gamal Elghazaly, Raphaël Frank\\
Interdisciplinary Centre for Security, Reliability and Trust (SnT), University of Luxembourg\\
29 Avenue John F. Kennedy, L-1855 Luxembourg, Luxembourg\\
{\tt\small firstname.lastname@uni.lu}
}

\begin{document}
\maketitle
\input{sec/0_abstract}

\begin{IEEEkeywords}
Autonomous driving, Scenario engineering, Synthetic data, Large language model (LLM)
\end{IEEEkeywords}

\input{sec/1_intro}
\input{sec/2_related_works}

\input{sec/3_method}

\input{sec/5_results}

\input{sec/6_conclusion}

{
    \small
    \bibliographystyle{IEEEtran}
    \bibliography{main}
}
\newpage
\onecolumn

\end{document}

%% file: sec/0_abstract.tex
\begin{abstract}

Rare and challenging driving scenarios are critical for autonomous vehicle development. Since they are difficult to encounter, simulating or generating them using generative models is a popular approach. Following previous efforts to structure driving scenario representations in a layer model, we propose a structured five-layer model to improve the evaluation and generation of rare scenarios. We use this model alongside large foundational models to generate new driving scenarios using a data augmentation strategy. Unlike previous representations, our structure introduces subclasses and characteristics for every agent of the scenario, allowing us to compare them using an embedding specific to our layer-model. We study and adapt two metrics to evaluate the relevance of a synthetic dataset in the context of a structured representation: the diversity score estimates how different the scenarios of a dataset are from one another, while the originality score calculates how similar a synthetic dataset is from a real reference set. This paper showcases both metrics in different generation setup, as well as a qualitative evaluation of synthetic videos generated from structured scenario descriptions. The code and extended results can be found at \href{https://github.com/Valgiz/5LMSG}{https://github.com/Valgiz/5LMSG}.

\end{abstract}




%% file: sec/1_intro.tex
\section{Introduction}
\label{sec:intro}

Edge Cases, Corner Cases, or Out of Distribution Scenarios are different names that generally refer to the most rare, challenging, or risky driving scenarios. They are very valuable for autonomous vehicle research, as they may be very challenging for learning-based systems~\cite{elhafsi2023semantic}, but they are also scarce among driving datasets by definition. Thus, a common workaround for researchers is to rely on simulations~\cite{dosovitskiy2017carla} and the generation of synthetic scenarios to acquire additional data~\cite{gao2025foundation, chang2024llmscenario}. Since definitions and names can vary across the literature~\cite{10252672}, this paper uses the term Edge Cases (ECs), to refer to rare and unexpected scenarios. They may consist of objects, agents, or behaviors that would not be expected to be found in a \textit{normal} everyday situtation. We specifically chose not to focus on the risky aspect of safety-critical scenarios, but more on the rarity aspect. Despite this limitation, knowing if a scenario can be considered rare is subjective, which reduces the specificity of this definition. Thus, our work focuses on scenarios that are rare relative to a reference set of known scenarios. What makes a scenario an EC will thus depend on that reference.

Our goal is to generate synthetic visual data of driving scenarios using generative AI that rely on World Foundation Models (WFM)~\cite{wang2024drivedreamer, veo3}. However, the variety of possible scenarios is extremely high. Before we generate visual data, we first focus on creating the content of ECs. In order to create these scenario descriptions, we rely on the vast general knowledge of Large Language Models (LLM) by guiding their capabilities to generate new scenario ideas. Following previous work on LLM-based driving scenario generation~\cite{aasi2024generating, chang2024llmscenario, gao2025foundation}, the challenge of this approach lies, not only in the proposed generation strategy, but also in the evaluation method and the scenario format. Both evaluation and structure depend on the use case and the type of scenario we aim to generate~\cite{chang2024llmscenario}. We propose a new format for textual representation of scenarios as a structured layered model~\cite{scholtes20216}, as well as an evaluation strategy tailored to this representation. Since physical realism can be challenging to measure from textual descriptions alone, we choose to generate synthetic scenarios based on existing real scenes to provide stronger chances that our scenarios will be coherent. Our generation strategy consists of augmenting a real dataset by editing targeted components of each scenario, and evaluating how our synthetic scenes differ from the original in order to grasp whether or not they constitute ECs. Rather than retraining a model with specific driving data, we study a prompting strategy on large pre-trained models. Our main contribution in this paper consists of the following:

\begin{enumerate}
    \item Proposing a structured 5-layer model for driving scenario representation.
    \item Proposing a new synthetic scenario generation strategy for driving dataset augmentation.
    \item Studying new text-based evaluation methods for driving scenarios. 
\end{enumerate}

The rest of the paper is organized as follows. Section~\ref{sec:related_works} covers related works in synthetic driving data generation. In Section~\ref{sec:method}, we explain our scenario representation, our generation strategy, and our evaluation metrics. Finally, we discuss our experimental results in Section~\ref{sec:results} before concluding this paper in Section~\ref{sec:conclusion}.

%% file: sec/2_related_works.tex
\section{Related works}
\label{sec:related_works}

\subsection{Conditioned video generation}

Conditional video generation has emerged as a prominent direction in visual generative modelling. Recent advances have shown substantial improvements both in visual fidelity and temporal coherence. These advances are particularly relevant in the context of autonomous driving, mainly in simulation, training, and validation, aiming to produce high-fidelity, temporally coherent, and controllable driving scenarios that capture real-world dynamics, including rare and critical situations. These methods can be categorized into (1) text-to-video~\cite{hu2022make,ho2022video,zhou2024storydiffusion}, (2) image-to-video~\cite{ni2023conditional,hu2022make,blattmann2023stable}, or (3) video-to-video~\cite{wu2024fairy, liang2024flowvid}. 
Recent research in video generation for autonomous driving has emphasized controllable and conditioned generation using both visual and semantic cues. Approaches based on diffusion transformers—such as Cosmos-Transfer1—enable realistic and spatiotemporally coherent videos by conditioning on segmentation maps, depth data, and text prompts relating to driving context (e.g., weather, traffic situations)~\cite{fu2025llm, gao2024vista, blattmann2023stable}. Models like DriveDreamer and GEM integrate high-level planning with text or action commands, structuring the generation process for “world models” that underpin robust simulation-to-reality transfer~\cite{wang2024drivedreamer, hassan2024gem, gao2024vista}. These methods are increasingly leveraging foundation models and Vision-Language Models (VLMs) to bridge simulation with natural scene understanding, allowing scenario synthesis from abstract or rare prompts (e.g., “child crossing in fog”)~\cite{fu2025llm, hu2023gaia1generativeworldmodel, chang2025seeing}. However, fine-tuning for specific domains has revealed a trade-off: while visual quality and fidelity often improve, some loss of dynamic semantic accuracy—such as vehicle-pedestrian interactions—can occur if the balance of objectives is not carefully maintained~\cite{chang2025seeing}.

\subsection{Synthetic datasets for autonomous driving}

Synthetic datasets remain crucial for scalable autonomous driving research, allowing exhaustive exploration of rare or hazardous scenarios~\cite{fu2025llm, dosovitskiy2017carla}. Platforms like CARLA~\cite{dosovitskiy2017carla} and datasets like nuScenes~\cite{caesar2020nuscenes} are extended by VLM-based generative pipelines, supporting both low-level (e.g., pixel-level augmentations, weather changes) and high-level (e.g., event-driven, safety-critical) scenario synthesis~\cite{fu2025llm, gao2024vista, wang2024drivedreamer}. Recent advances combine programmatic scene specification (via LLM-generated code) with conditional video generation to efficiently populate training and testing datasets, targeting underrepresented edge conditions~\cite{fu2025llm, aasi2024generating}. Data augmentation methods range from traditional transforms to LLM-guided scenario recombination and adversarial perturbation~\cite{fu2025llm, russell2025gaia}. Importantly, these advances have enabled a significant boost in perception and planning model robustness, establishing synthetic datasets as a foundational pillar of modern AV system benchmarking~\cite{blattmann2023stable,hu2023gaia1generativeworldmodel}.

\subsection{Evaluation of generated visual scenarios}

Evaluating generated driving videos for autonomous vehicles demands a multifaceted approach. Classical perceptual metrics like Fréchet Inception Distance (FID) and Fréchet Video Distance (FVD)~\cite{heusel2017gans, unterthiner2018towards} remain standard for visual realism, but recent studies stress the need for semantic metrics tailored to driving: object-level accuracy (e.g., segmentation/instance tracking), event coverage (e.g., collision, rare maneuver occurrence), and scenario diversity~\cite{chang2025seeing,gao2024vista,fu2025llm}. Human-in-the-loop and AI-judge protocols (e.g., GPT-4o forced-choice or scenario-based QA) are increasingly employed to supplement numerical scores with qualitative validation~\cite{chang2025seeing,blattmann2023align}. Furthermore, “scenario-based evaluation”—requiring downstream perception, planning, and policy modules to operate safely and robustly on generated samples has emerged as a key paradigm~\cite{chang2025seeing}. Catastrophic forgetting and reward misspecification during fine-tuning are studied for their effects on both surface realism and behavioral fidelity, underscoring the subtle interplay between visual, semantic, and operational evaluation criteria~\cite{chang2025seeing,parisi2019continual}.

%% file: sec/3_method.tex
\section{Method}

\label{sec:method}

\begin{figure*}[ht!]
    \centering
    \includegraphics[width=0.89\linewidth]{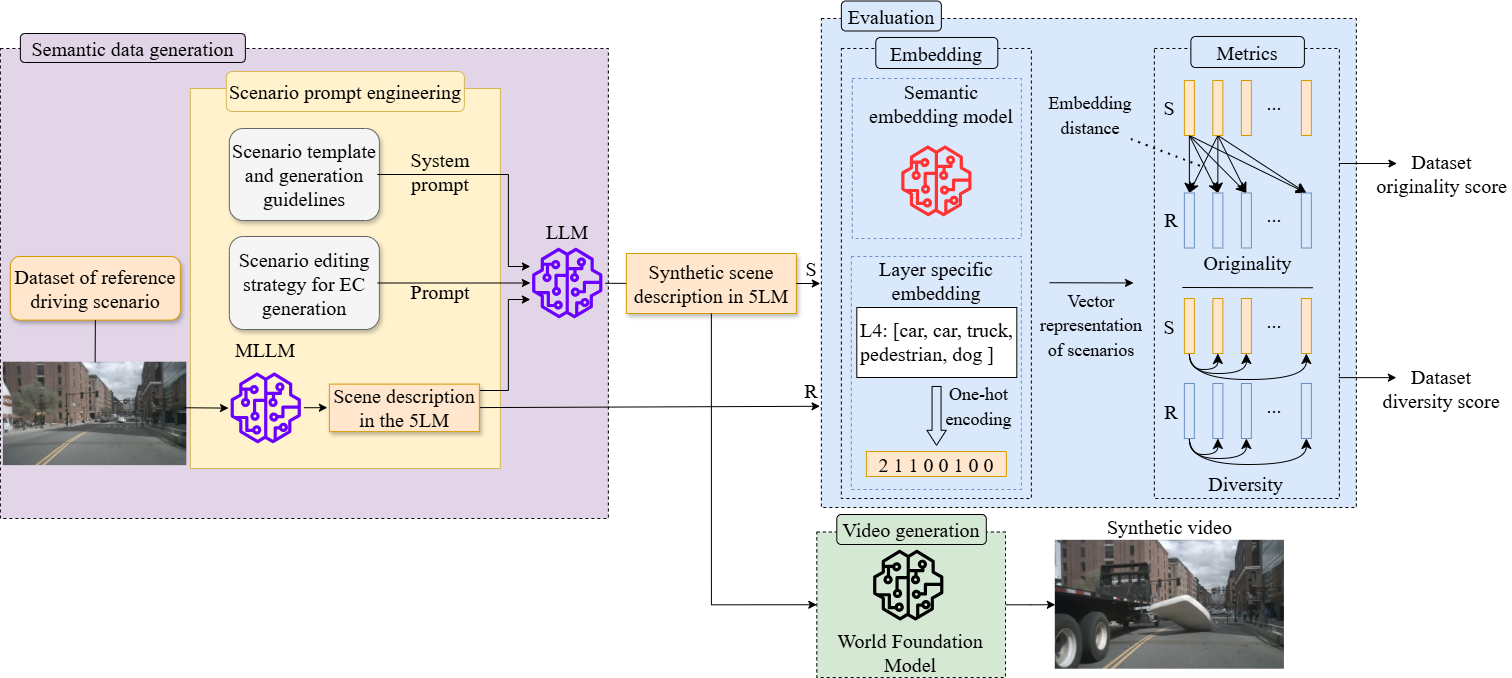}
    \caption{Proposed method for generating and evaluating new diverse scenario from real scenes. Our generation strategy relies on editing real scenarios after representing in a 5-layer model.}
    \vspace{-0.3cm}
    \label{fig:method}
\end{figure*} 

Since the realism of a textual description provided by a language model is hard to evaluate, we choose to rely on a data augmentation strategy that takes existing real scenarios and modifies specific components to create ECs. This method at least guarantees that a large part of the scene is already grounded in reality and provides a very specific context for the LLM. Our evaluation strategy relies on representing textual descriptions as vector embeddings so that we can calculate a similarity score between two scenarios (Fig.~\ref{fig:method}). 
Section~\ref{sec:structure} focuses on our scenario representation while Section~\ref{sec:gen_strat} details our generation strategy for driving scenarios. The embeddings, as well as the associated metrics, are described in detail in Section~\ref{sec:evaluation}.

\subsection{Structured driving scenarios for explainable scenario generation with an LLM} 
\label{sec:structure}

When directly prompting an LLM to generate loosely defined concepts like ECs, we do not know if the model really understands the output we expect.
More generally, a structured formatting of the expected scenarios can help an LLM generate higher-quality and diverse responses. We rely on a 5-layer model (5LM) based on the 6LM proposed by \textit{Scholtes et al.}~\cite{scholtes20216} that decomposes a driving scenario into five different layers, namely (a) road structures, (b) structures surrounding the road, (c) temporary changes to the first two layers, (d) dynamic objects, and finally (e) environmental conditions. This model not only standardizes the representation of a driving scenario, but also allows us to study prompting strategies that target very specific elements of the scenario (Fig.~\ref{fig:Editing_strat}). In their model, \textit{Scholtes et al.} use a sixth layer for digital information, such as V2X messages, which are out of scope of the present work. Since our textual synthetic scenarios will be used to generate visual data, we ignore the digital layer which is not visible.
Initially, our proposed structure contains only a string of text for each layer, producing an \textit{unstructured 5-layer model}. 
Based on this model, we define a driving scenario $s$ to be the concatenation of a vector of textual descriptions as: 
\begin{equation}
 s= \bigoplus_{k = 1}^{5} L_k,   
\end{equation}
\noindent where \(L_k\) is the textual description of the content of layer $k$, $\forall k = \{1, \, 2, \, \cdots, \, 5 \}$. 
Eventually, this scenario description is further improved by introducing more details about the content of each layer while constraining the model to fill a template for each layer. 
This template creates classes of objects expected in the layer intended for modifications and indicates the minimum characteristics expected for each object. We refer to this model as a \textit{structured 5-layer model} (Fig.~\ref{fig:structure_5LM}).


\subsection{Scenario generation strategies}
\label{sec:gen_strat}

\subsubsection{Reference scenarios}
The method we propose is applicable to textual representations of driving scenarios and not to visual representations such as videos. However, driving datasets consist mostly of image data.
Descriptions in common datasets such as NuScenes~\cite{caesar2020nuscenes} specify only some of the agents involved without detailed information on the infrastructures or the environment. In order to acquire real reference scenes from which we will generate our scenarios, we use a Multimodal Large Language Model (MLLM) to analyze visual data from a driving dataset and format it according to our model (Fig.~\ref{fig:method}). For this paper, the nuScenes dataset~\cite{caesar2020nuscenes} and the Gemini 2.5 Pro MLLM~\cite{comanici2025gemini} were used, as they are a popular SotA dataset and model, respectively.

\subsubsection{Unstructured and non-specific layer editing}

\begin{figure}[t]
\centering
\includegraphics[width=0.8\linewidth]{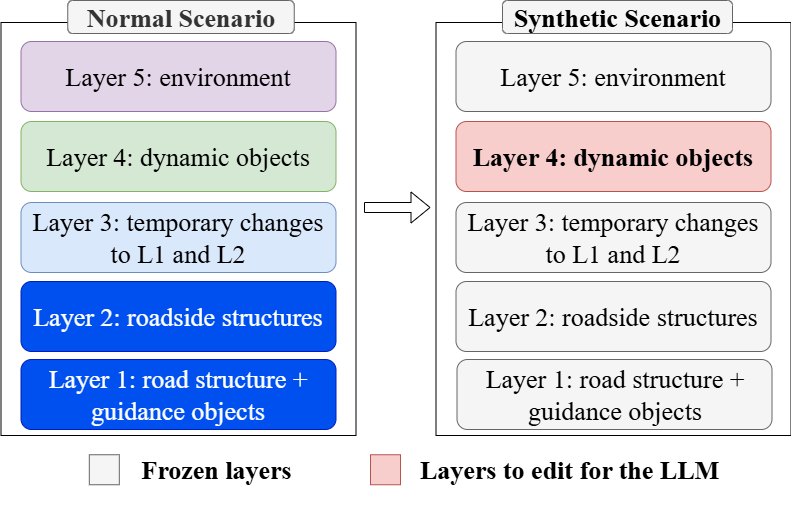}
\vspace{-0.3cm}
\caption{Application of the layer model to scenario generation. The LLM is prompted with editing specific layers at a time to increase our control over the generation process. Similarly, the evaluation is also done layer per layer for increased scene understanding}
\label{fig:Editing_strat}
\vspace{-0.4cm}
\end{figure}

Our initial editing strategy consists of prompting the model to edit a specific layer of a real scenario while leaving the other layers unchanged (Fig.~\ref{fig:Editing_strat}). We build our prompt by first giving the model a system prompt that explains its role, provides guidelines on how to understand the 5-layers model and explain the concept of EC. At this stage, the scene is decomposed into 5 layers that are just described by unstructured text. Whenever any specific layer \(L_i\) is modified, the model is simply asked to modify the given layer:
%
\task{Please only modify the layer specified in the prompt to generate an Edge Case and change nothing in the other layers (MOST IMPORTANT)}

\subsubsection{Structured layer-wise editing} \label{layer_specific_task}
\begin{figure*}[t]
    \centering
    \includegraphics[width=0.95\linewidth]{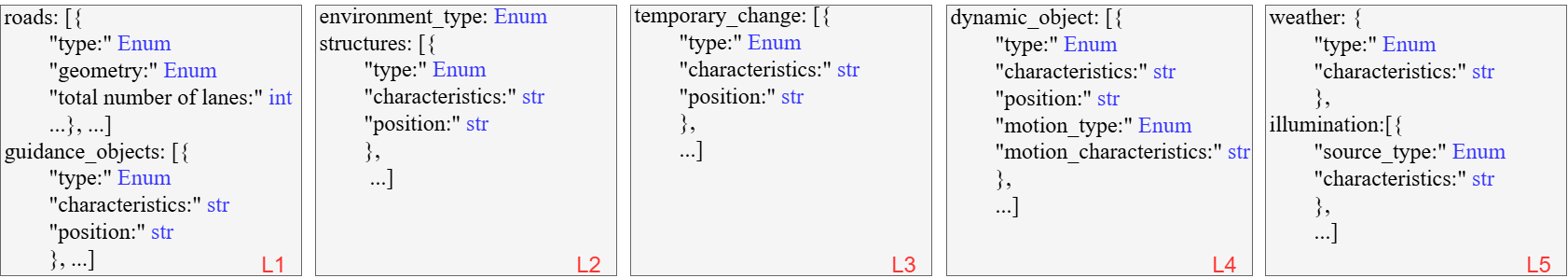}
    \caption{Snapshot of the structured 5-layer model representation. The Enum fields correspond to a list of available choice for the model, while the str fields are more free section for the model to specify additional information. Some fields like structures (L2), dynamic objects (L4) or illumination (L5) expect a list of all relevant object fitting that category within the scene.}
    \vspace{-0.3cm}
    \label{fig:structure_5LM}
\end{figure*}

Our second approach consists of providing the model with additional structure within each layer. Using the functionality of the Gemini API, we can force the model output to follow a given JSON template (Fig.~\ref{fig:structure_5LM}). Each layer contains specific classes of expected objects and their attributes. Any additional textual details about these objects will be placed in a \textit{'Characteristics'} attribute, which we can use for text-based evaluation, like the previous strategy. This allows us to define layer-wise metrics (see Section~\ref{sec:evaluation}) and to introduce a different input prompt well-adapted for each layer.
Unlike the previous subsection, when the model is tasked with editing a layer, such as \(L_4\), it will receive an additional task specific to that layer: 
\task{Turn this scenario into an Edge Case by modifying only the layer L4 from the input. You should either: - Modify existing dynamic objects, or add new ones with rare and/or challenging characteristics. Look for object that do not belong in such a scenario. - Modify the motion of existing dynamic objects, or add new objects with unique and challenging motion. You may do both if needed, but focus on either the characteristics or the motion of the objects when generating a scenario.}

\subsection{Evaluation metrics}
\label{sec:evaluation}

In order to evaluate the proposed method for generating scenarios, we have to identify relevant evaluation metrics. The final goal of this method would be to generate visual data for EC driving scenarios. However, existing evaluation metrics for synthetic driving data would mostly evaluate the visual performance of the final method used to render these scenarios. We aim to evaluate the relevance of our synthetic scenarios at the textual description level before video generation. Considering that we are looking for ECs, we need to evaluate whether the generated scenario is considered a \textit{rare} and \textit{unexpected} event. Among existing text-based evaluation metrics, \textit{Aasi et al.} proposed two metrics to quantify Diversity and Out-of-Domainness (OOD)~\cite{aasi2024generating}, based on the calculation of the semantic similarity between two given scenarios. Following a similar strategy, we developed two sets of evaluations. The semantic embedding-based metrics use the method from \textit{Aasi et al.}, changing the distance measurement formulas from a sample to a dataset in both of their metrics. It uses the Gemini-embedding-001 model~\cite{lee2025gemini} to calculate semantic similarities within the \textit{unstructured 5LM} (Fig.~\ref{fig:metrics}). Our second approach relies on the \textit{structured 5LM} to calculate the similarity of the scenarios based on a structure-specific vector representation. Although it is subjective to precisely define what can be considered an EC, these metrics allow us to compare our synthetic data with a real reference, in order to evaluate if they are an EC in regard to that dataset. In both cases, we calculate the similarity between two embeddings using the cosine similarity:

\begin{equation}
    \delta(e_i, e_j) = \frac{e_i^T.e_j}{||e_i|| \, ||e_j||},
\end{equation}

\noindent where $e_i$ and $e_j$ are the embeddings of two scenarios that are being compared. If $e_i$ and $e_j$ are identical, the cosine value will be 1. The minimum similarity value will be 0 if $e_i$ and $e_j$ are perpendicular, since $e_i^T.e_j = 0$.

\label{sec:semantic_eval}
Our initial metrics are based on semantic embedding distances between texts. Our goal is to use an embedding model to obtain a quantitative evaluation of how similar our synthetic scenarios are to the reference samples. From that we can quantify how similar our synthetic dataset is to the reference dataset. However, unlike \textit{Aasi et al.}~\cite{aasi2024generating} who directly compare the distances between the full description of their scenes and the reference, we rely on the unstructured 5LM to compare each layer of the scenes separately in order to achieve a more precise understanding of the scene's originality. Let \(S_{gen} = \{s_{g,1},..., s_{g,M}\}\) and \(S_{ref} = \{s_{r,1},..., s_{r,N}\}\) be the set of generated synthetic scenes and real reference scenes respectively. Each scene \(s\) is represented in the 5LM as the texts \(s_k, \forall k \in[1,5]\) that represent the descriptions of the corresponding layers. For each layer, we compute the embeddings of both sets \(E_{gen,k} = \{e_{g,1,k},e_{g,2,k}..., e_{g,M,k}\}\) and \(E_{ref,k} = \{e_{r,1,k},e_{r,2,k}..., e_{r,N,k}\}\), where \(e_{g,i,k}\) is the embedding of the \(k\)-th layer of the \(i\)-th generated scene \(s_{g,i} \in S_{gen}\). In this section, the embeddings are computed using Gemini-embedding-001 as it ranks first on several embedding model leaderboards~\cite{lee2025gemini} and has been trained on the same data as Gemini-2.5, which we use as the LLM and MLLM of our pipeline.

\subsubsection{Originality}

The Originality score \(O_{g,j,k}^r\) of a sample \(s_{g,j}\) from a reference set \(S_{ref}\) for layer \(L_k\) is calculated as:

\begin{equation}
    O_{g,j,k}^{r}=\underset{i\in\{1,...,N\}}{\text{max}} \delta(e_{r,i,k}, e_{g,j,k})
\end{equation}

Unlike us, \textit{Aasi et al.} calculate the OOD score $O_{g,j}^{r}$ between \(s_{g,j}\) and \(S_{ref}\) using the sample \(e_{r,i}\) with \textbf{minimum} similarity instead of \textbf{maximum}. Their metrics measure the similarity from the generated sample to the least similar (i.e. the most different) sample within the reference set, while we look at the reference sample that is the closest from our synthetic scenario. We found this to be more informative than the original method, which cannot distinguish if the generated sample is already present in the reference dataset. The final originality scores of the generated layers \(L_k\) are then given by the average originality of every new scene with the original set: 

\begin{equation}
    O_{g,k}^{r}=\frac{1}{M}\overset{M}{\underset{j=1}{\sum}} O_{g,j,k}^{r}
    \label{eq:OOD}
\end{equation}
\subsubsection{Diversity}
\begin{figure}[t]
    \centering
    \includegraphics[width=0.9\linewidth]{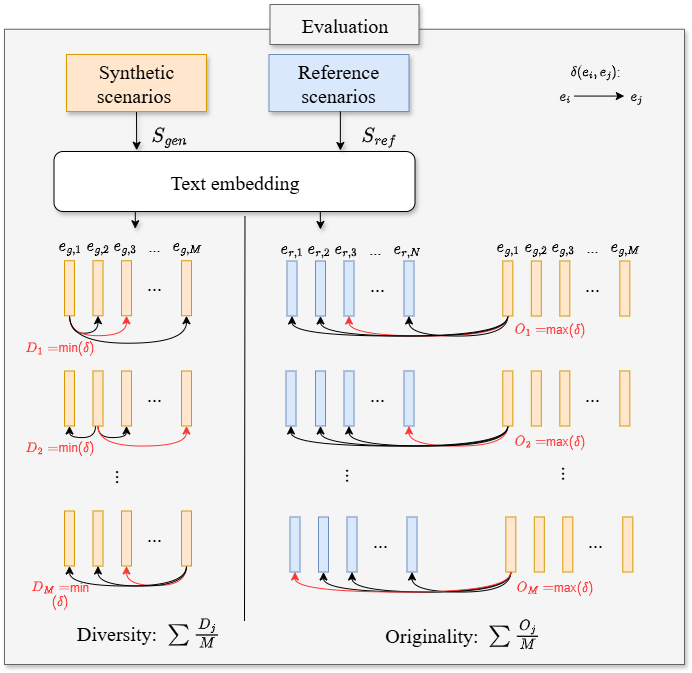}
    \caption{Diversity and Originality are calculated using the semantic distances between the embeddings of the generated scenarios \(S_o\) and the reference scenarios \(S_b\).}
    \vspace{-0.3cm}
    \label{fig:metrics}
\end{figure}

The Originality metric was designed to quantify how different our scenarios are from the reference, since ECs are defined to be unknown events. Similarly, the diversity metric aims to identify whether the scenarios within a set are similar to each other by evaluating the similarity between each sample (Fig.~\ref{fig:metrics}). Unlike the previous metric which provides a single score to compare synthetic and reference scene, this metric provide a score for each set separately.
\begin{equation}
    D_{g,j,k}=\underset{\underset{ {j'\ne j}}{j'  \in\{1,...,M\}}}{\text{min}} \delta(e_{g,j',k}, e_{g,j,k} )
\end{equation}
\begin{equation}
    D_{g,k}=\frac{1}{M}\overset{M}{\underset{j=1}{\sum}} D_{g,j,k}
    \label{eq:div}
\end{equation}

Like the previous metric, we differ from \textit{Aasi et al.} not just with the layer model, but because they calculate diversity as the mean of the \textbf{maximum} similarity between samples. However, we found this version to be too penalizing for larger datasets, as increasing the number of samples in a dataset could drastically reduce its diversity, even if it covers a wider range of scenarios. Especially considering that our objective is to generate many new scenarios from a small amount of reference. We will refer to both of these metrics as Diversity(max) and Diversity(min), for the versions of \textit{Aasi et al.} and ours respectively. 


\subsubsection{Layer-wise evaluation}
\label{sec:layer-wise-eval}
Each layer of the 5LM represents a fundamentally different component of a scenario based on its property, its behavior, or its interaction with the ego vehicle. Although measuring metrics separately for each layer provides some insight into the relevance of the new scenes, our structured model allows us to define a layer-wise evaluation strategy to account for the specificity of each layer.
In the structured 5LM (Fig.~\ref{fig:structure_5LM}), each component of a layer has a mandatory discrete \textit{type} parameter with predefined and limited possible values. Instead of using the pre-trained embedding model, we rely on this representation to define a structure-specific vector for the layer of a scenario.

For any layer $L_k$ with $k\in \{1,2,...,5\}$ containing a list of objects with a \textit{type} parameter, with $C_k$ possible values. $L_k$ is represented as the vector $V_k$ of dimension $C_k$. It is defined as the count of objects within $L_k$ in each category.
Take a scene s where \(L_4 = [\)
\{'type': 'vehicle', ...\}, \{'type': 'inanimate object', ...\}, \{'type': 'vehicle', ...\}, \{'type': 'pedestrian', ...\}$]$ and the available categories are [vehicle, cyclist, pedestrian, animal, inanimate object, other]. $C_k = 6$ and we can embed $L_4$ as the vector:
\(V_4(s) = \begin{pmatrix}
    2 & 0 & 1 & 0 & 1 & 0
\end{pmatrix}^T\).

This vector can be considered as an embedding of the layer within the type space associated with the expected objects within that layer. Like Section~\ref{sec:semantic_eval}, we can also use that vector to calculate the distance between the distribution of objects present in the layer of each scene. It gives us a new version of diversity and originality which we call Component Diversity (CD) and Component Originality (CO).
However, the very small size of these embedding spaces (less than ten dimensions) also reduces the relevance of the diversity metric, as samples are much more likely to reach a similarity score of 0 or 1. Thus, the diversity(min) that we used for previous evaluation would easily stay at 0 in generation experiments, making even 1 outlier too impactful on the Diversity score. Similarly, the diversity(max) of \textit{Aasi et al.} would mostly remain at 1. To find a better metric, we compromised between our approach and \textit{Aasi et al.} by using another diversity score using the average similarity of a sample with every other sample in the same set: 
\begin{equation}
      D_{g,j,k}=\frac{1}{M-1} \underset{\underset{ {j'\ne j}}{j'  \in\{1,...,M\}}}{\sum} \delta(V_k(s_{g,j'}), V_k(s_{g,j})) 
\end{equation}

%% file: sec/5_results.tex
\section{Results and Discussion}
\label{sec:results}

\subsection{Experimental setup}
Our strategy was implemented on the nuScenes dataset\cite{caesar2020nuscenes}. Specifically, we used the small sample nuScenes-mini containing 10 scenes of twenty seconds. Each video was provided to the MLLM Gemini-2.5-Pro, which was tasked with providing a textual description following our template (see Section~\ref{sec:structure}). Two types of description were generated, corresponding to the structured and unstructured layer model described in Section~\ref{sec:method}. 
For each experiment, we used all 10 scenes to generate new scenarios and created 10 new scenarios for each layer, giving us 500 generated descriptions for each experiment. The LLM used is also Gemini-2.5-pro, while the embedding model used for the evaluation metrics is Gemini-embedding-001. The model heat is kept at the default value of 1 for all experiments.

\subsection{Results}

\subsubsection{Context and Metrics}
In Table~\ref{tab:exp_1}, the originality and diversity are calculated for each layer and using both the maximum and minimum semantic similarity values. The OOD-ness with minimum similarity (O(min)) and the diversity with maximum similarity (D(max)) correspond to the metrics presented by \textit{Aasi et al.}~\cite{aasi2024generating}, while the other two are defined in Equations~\ref{eq:OOD} and~\ref{eq:div}. All samples in this experiment are generated in the unstructured 5LM. In addition, this experiment evaluates the impact of using the same context to generate scenarios. In one case, the model is called 10 times independently for each layer \(L_k\) to generate 1 new scenario by editing \(L_k\). In the other case, the model is directly asked once to provide 10 new scenarios. Overall, it appears that the context's influence is very small, with slightly better results when using the same context. We can also notice that the embedding space of Gemini-embedding-001 puts our textual descriptions close to each other, since the lowest diversity score is only at 0.78. For all of our experiments, the embedding distance between two samples using this model never reached a value lower than 0.7. In addition to the previously mentioned issues of the metrics of \textit{Aasi et al.} (Section.~\ref{sec:semantic_eval}), our metric identify a difference between same and different context more often. The values of D(max) in particular always stays very close to 1 making it hard to use. Thus, we chose to keep our versions of the metrics (i.e. O(max) and D(min)).

\begin{table}[h!]
    \centering
        \caption{For each layer, synthetic scenes were generated either by making 10 independent queries to the model ("$\neq$" context), or all 10  scenes where asked in the same prompt("$=$" context).}
        \fontsize{9pt}{11pt}\selectfont
    \begin{tabular}{c|p{0.4cm}p{0.4cm}p{0.4cm}p{0.4cm}p{0.4cm}|c|c}
         \hline
         \hline
         Metric &  L1 & L2 & L3 & L4 & L5 & Total & Context\\
         \hline
         \hline
         \textit{Aasi}: O(min)& 0.86 & 0.84 & 0.82 & 0.84 & 0.80 & 0.89 & $\neq$ \\ & 0.86 & 0.84 & \textbf{0.81} & 0.84 & \textbf{0.79} & 0.89 & $=$\\
         \hline
         Ours: O(max)& 0.91 & 0.89 & 0.86 & 0.90 & 0.85 & \textbf{0.97} & $\neq$\\ & 0.91 & \textbf{0.88} & \textbf{0.85} & \textbf{0.89} & \textbf{0.83} & 0.98 & $=$\\
        \hline
        Ours: D(min) & 0.89 & 0.88 & 0.78 & 0.87 & 0.83 & 0.90 & $\neq$\\ & \textbf{0.88} & \textbf{0.85} & 0.78 & \textbf{0.84} & \textbf{0.82} & 0.90 & $=$\\
        \hline
        \textit{Aasi}: D(max) & 0.99 & 0.99 & 0.98 & 0.99 & 0.99 & 0.99 & $\neq$\\ & 0.99 & 0.99 & 0.98 & 0.99 & 0.99 & \textbf{0.97} & $=$\\
        \hline
        \hline
    \end{tabular}
    \label{tab:exp_1}
\end{table}

\subsubsection{Scenario Structure}

In Table~\ref{tab:exp_2}, we compare three different levels of scenario structure to understand the impact of structured scenarios: (a) The unstructured prompt uses a 5-layer model, where the content of each layer is just a string of text. The model is given guidelines on what type of information should go in which layer, but the generation process inside a layer is left free. (b) The 'Soft' structure adds more detail from the previous prompt by specifying a few components the model must mention, as well as some of their characteristics. It adds a specific task prompt depending on which layer is being edited (see Section~\ref{layer_specific_task}). (c) The 'Hard' structure defines a template with classes for each component that specify the information that must be filled. Unlike the previous method, where the soft structured output is still being returned as plain text for each layer, we use Gemini's structured output functionality to force the model to reply with a JSON file following our template (Fig.~\ref{fig:structure_5LM}).
    
    Table~\ref{tab:exp_2} shows that all synthetic datasets have a better diversity than the reference scenarios (0.88, 0.81 and 0.88 on synthetic layer 1 against 0.92 on reference). While the soft structure always performs better than the unstructured prompt, the hard structure appears to perform worse than the soft structure, and sometime worse than the unstructured prompt. Although adding more guidance for the model on required information helped its performance, the hard model may suffer from a non-adapted evaluation method. A large part of the text for the hard structure corresponds to the definition of the template, which may impact its diversity, since a large part of its text is identical between samples.  


\begin{table}[h!]
    \centering
        \caption{Results of the unstructured 5LM (U), soft (S) and hard (H) structured 5LM. The reference text (R) are the real scenario from nuScenes.}
        \fontsize{9pt}{11pt}\selectfont
    \begin{tabular}{p{1.2cm}|p{1.2cm}|ccccc}
    \hline
    \hline
        \centering Metric & \centering Structure &  L1 & L2 & L3 & L4 & L5 \\
         \hline
         \hline
        & \centering U & 0.91 & 0.88 & 0.85 & 0.89 & 0.83\\
        \centering $O$ & \centering S & 0.87 & \textbf{0.84} & \textbf{0.83} & \textbf{0.84} & 0.82\\
        & \centering H & \textbf{0.86} & \textbf{0.84} & 0.84 & 0.86 & \textbf{0.81}\\
        \hline
        & \centering R & 0.92 & 0.92 & 0.89 & 0.92 & 0.96\\ 
        & \centering U & 0.88 & 0.85 & 0.78 & 0.84 & 0.82\\
        \centering $D$ & \centering S & \textbf{0.81} & \textbf{0.78} & \textbf{0.76} & \textbf{0.78} & \textbf{0.80}\\
        & \centering H & 0.88 & 0.83 & 0.83 & 0.86 & 0.81\\
        \hline
        \hline
    \end{tabular}
    \label{tab:exp_2}
\end{table}

\subsubsection{Layer wise Evaluation}

\begin{figure}
    \centering
    \includegraphics[width=\linewidth]{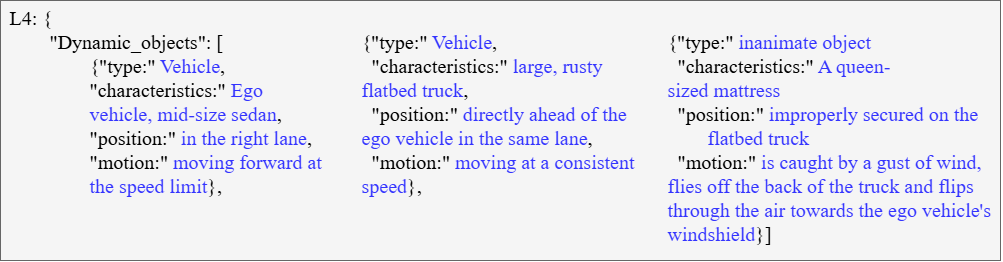}
    \caption{Example of an edited layer 4 from scene 1 of nuScenes in the \textit{structured 5LM}. The truck and mattress have been added to the scene.}
    \vspace{-0.6cm}
    \label{fig:L4_example}
\end{figure}


\begin{table*}[t!]
    \centering
        \caption{The object diversity calculates the diversity score as defined in Section~\ref{sec:semantic_eval} on the corresponding components of that layer using the text in their characteristics fields using Gemini-embedding-001.}
    \begin{tabular}{c|cc|cc|c|cc|cc}
         \hline
         \hline
        layer &  L1 && L2 && L3 & L4 && L5 \\
         \hline
         layer component & roads & guidance & environment & structure & object & object & motion & weather & illumination \\
         \hline
          Generated scenes & \textbf{0.86} & \textbf{0.89} & \textbf{0.74} & \textbf{0.85} & \textbf{0.79} & \textbf{0.71} & \textbf{0.76} & \textbf{0.79} & \textbf{0.83}\\
        Reference scenes& 0.91& 0.91 & 0.83 & 0.92 & 0.80 & 0.74 & 0.81 & 0.84 & 0.87\\
        \hline
        \hline
    \end{tabular}
    \label{tab:exp_3_char}

\end{table*}

\begin{table}[t!]
    \centering
        \caption{The CD and CO metrics use the vector defined in Section~\ref{sec:layer-wise-eval}.}
    \begin{tabular}{c|c|ccccc}
         \hline
         \hline
         & Metric &  L1 & L2 & L3 & L4 & L5 \\
         \hline
         \hline
         
        Generated
        Scenes& CO & 0.93 & 0.59 & NA  & 0.85 & 0.81\\
        & CD & 0.68 & \textbf{0.18} & NA & \textbf{0.63} & \textbf{0.50} \\
        \hline
        Reference Scenes& CD & \textbf{0.54} & 0.64 & NA & 0.91 & 0.52\\
        \hline
        \hline
    \end{tabular}
    \label{tab:exp_3}

\end{table}

The layer-wise evaluation is highly dependent on the structure used for the scenario, and especially the enumeration of the possible types of object, environments, or roads. In the results of Table~\ref{tab:exp_3}, the generated scenario CD mainly outperforms the reference (0.18 to 0.64 for L2) but occasionally underperforms (0.68 to 0.54 for L1). In addition to the definition of the layer's structure, we found that when generating modified layers, the LLM have a tendency to remove or group objects from the original scene to focus mostly on the new content (see Fig.~\ref{fig:L4_example}). The average number of dynamic objects per scene in layer 4 falls from 6.4 in the reference to 2.7 in the generated scenes. Either the model does not repeat objects, or group them together as one: 4 car objects can be grouped as: \textbf{\{'type': 'Vehicle', 'characteristic': 'A mix of sedans and SUVs.', 'position': 'In the opposite lane.'\}}. In addition to our layer-specific embedding, we also evaluate the diversity of some components by computing the embedding of the \textit{characteristics} field, which remains pure text, and calculating the Diversity (min) in Table~\ref{tab:exp_3_char}. For each value, the generated scenes outperform the reference by having a better diversity (i.e., a lower Diversity score) of characteristics among objects, but the gap varies greatly depending on the layer. Note that L3 was underexplored in our study. Since our \textit{structured 5LM} had too few categories for the 'type' field, we chose to leave it as free text which make the CO and CD not applicable to that layer.

\begin{figure*}[h!]
    \centering
    \begin{subfigure}[t]{0.28\textwidth}
    \includegraphics[width=\textwidth]{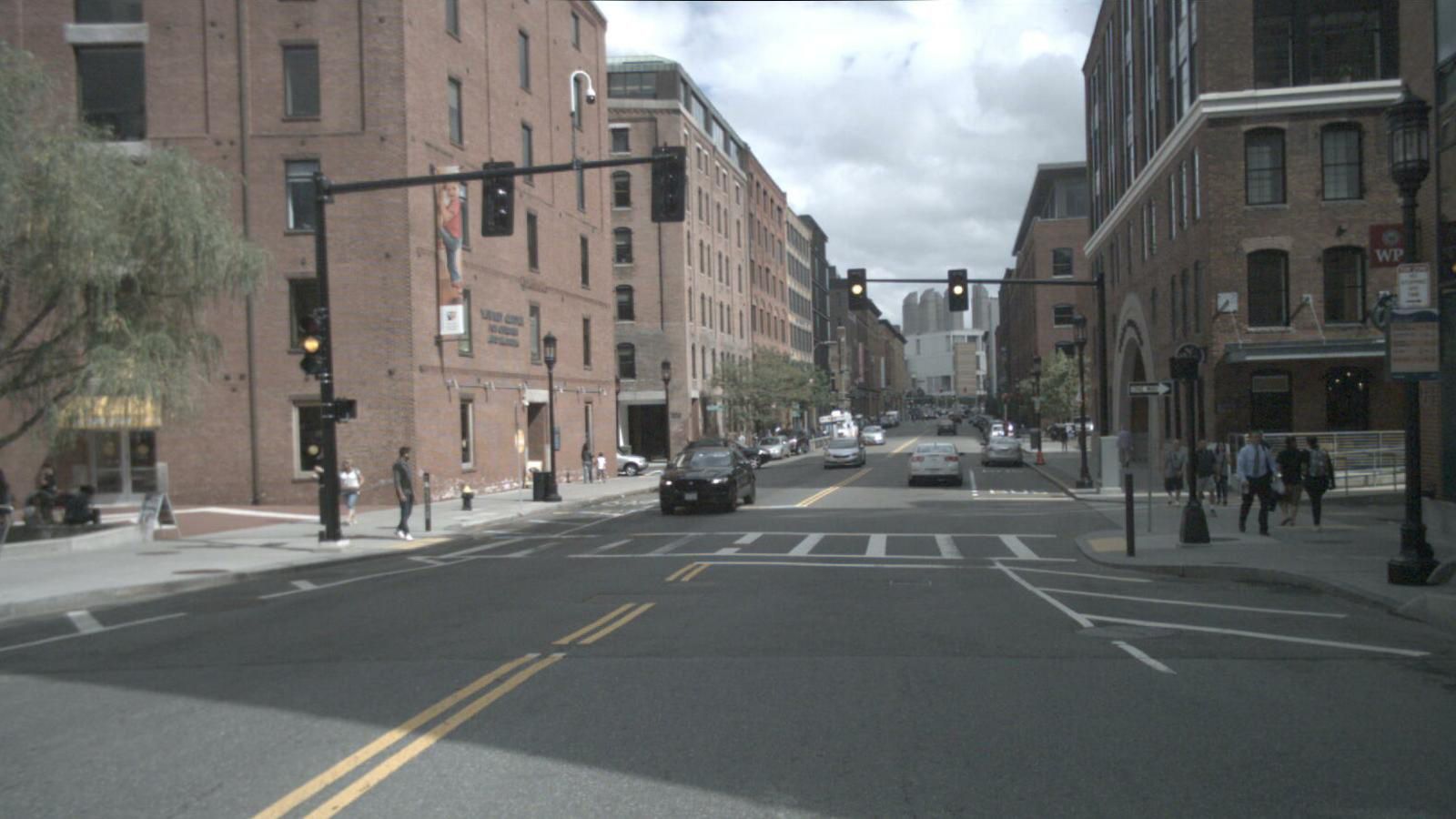}
    \caption{Front view camera of the scene 1 in the nuScenes dataset, at frame 1.}
    \label{subfig:ref_image}
    \end{subfigure}\hfill
    \begin{subfigure}[t]{0.28\textwidth}
    \includegraphics[width=\textwidth]{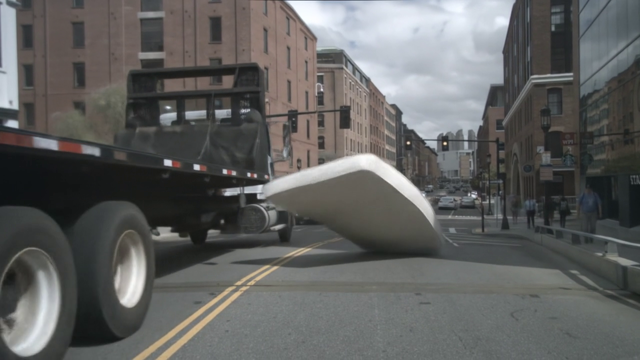}
    \caption{Video generated using the reference image~\ref{subfig:ref_image}, with a prompt specifying the edited layer from Fig.~\ref{fig:L4_example}, at 4 seconds.}
    \label{subfig:L4_guided}
    \end{subfigure}\hfill
    \begin{subfigure}[t]{0.28\textwidth}
    \includegraphics[width=\textwidth]{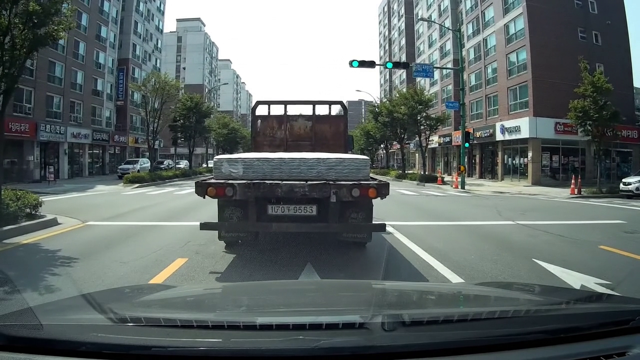}
    \caption{Video generated with the same prompt as Fig.~\ref{subfig:L4_guided} without image guidance, at 4 seconds.}
    \label{subfig:L4_non_guided}
    \end{subfigure}

    \vspace{-0.2cm}
    \caption{Screenshots of synthetic videos generation using the Veo3-preview WFM on our synthetic scenario shown in Fig.\ref{fig:L4_example}.}
    \label{fig:scene 1}
    \vspace{-0.2cm}
\end{figure*}

\subsubsection{Qualitative evaluation of video generation}

\begin{figure*}[t!]
    \centering
    \begin{tabular}{ccccc}
         \rotatebox{90}{\footnotesize  $\,\,\,\,\,\,\,\,\,\,\,\,\,\text{Scene 1}$} & \includegraphics[width=3.8cm]{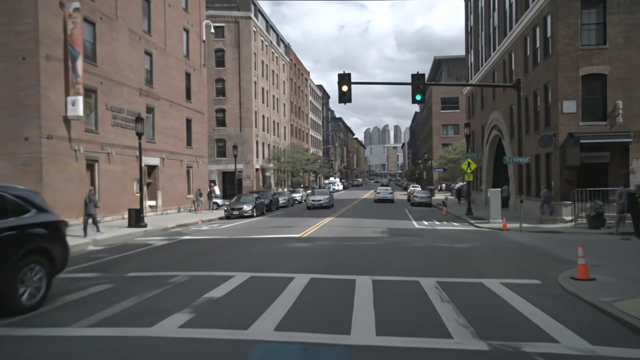} & \includegraphics[width=3.8cm]{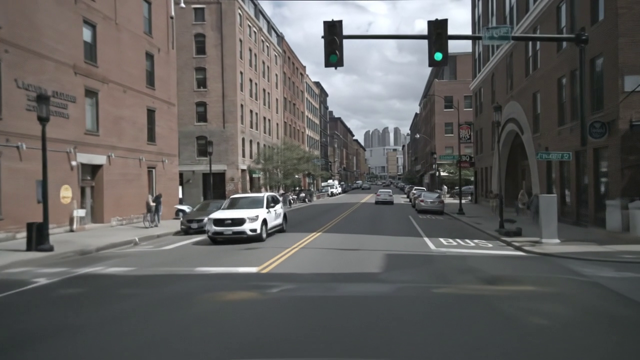} & \includegraphics[width=3.8cm]{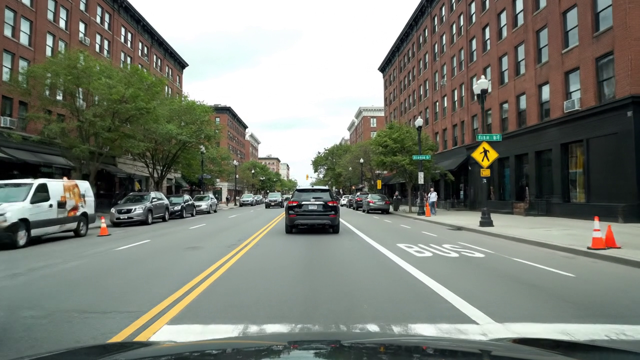} & \includegraphics[width=3.8cm]{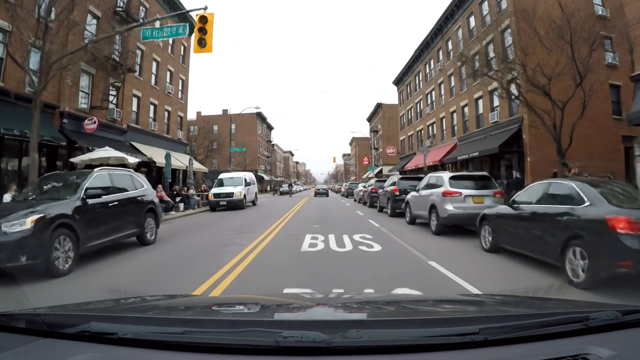}\\
        \rotatebox{90}{\footnotesize  $\,\,\,\,\,\,\,\,\,\,\,\,\,\text{Scene 3}$} & \includegraphics[width=3.8cm]{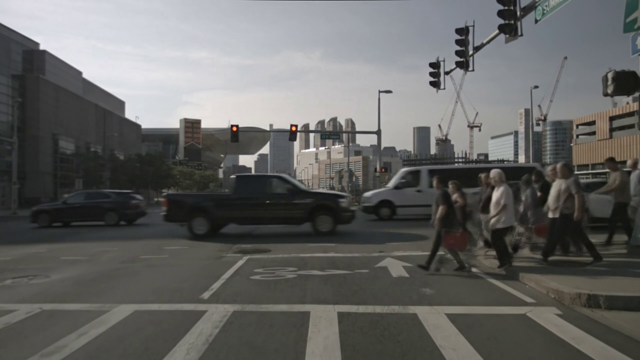} & \includegraphics[width=3.8cm]{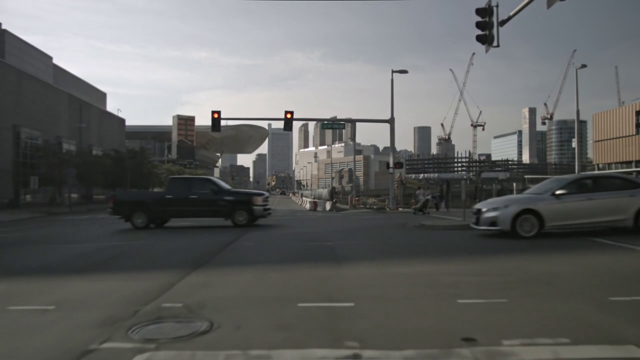} & \includegraphics[width=3.8cm]{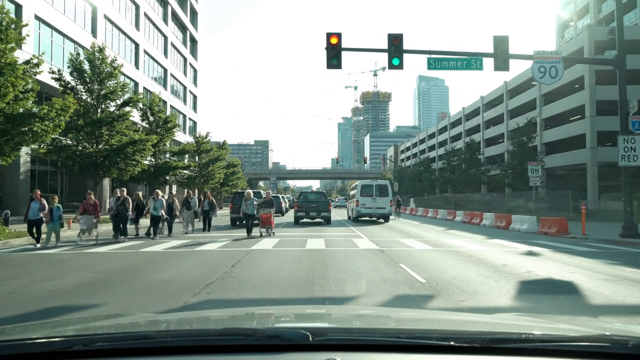} & \includegraphics[width=3.8cm]{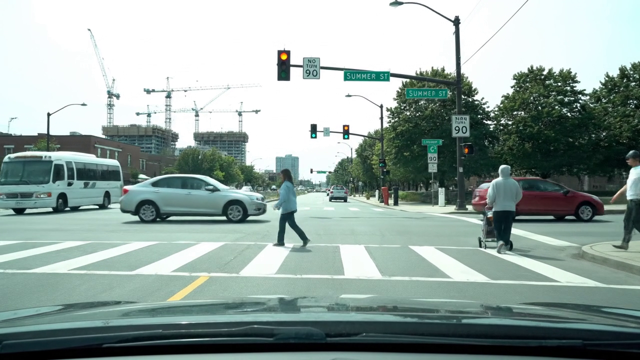} \\
        \rotatebox{90}{\footnotesize  $\,\,\,\,\,\,\,\,\,\,\,\,\,\text{Scene 10}$} & \includegraphics[width=3.8cm]{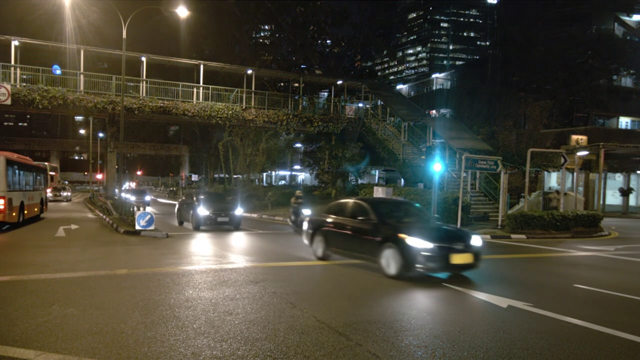} & \includegraphics[width=3.8cm]{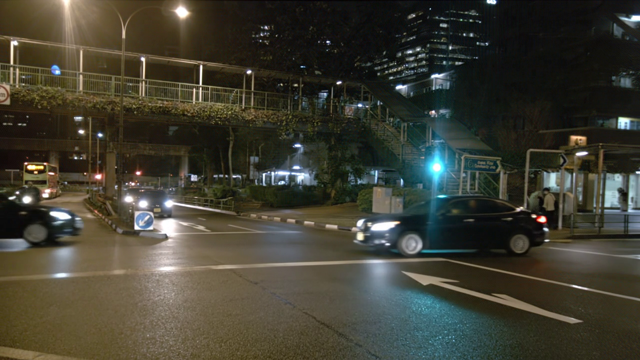} & \includegraphics[width=3.8cm]{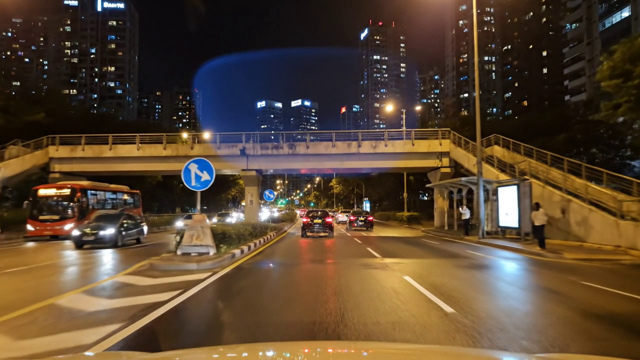} & \includegraphics[width=3.8cm]{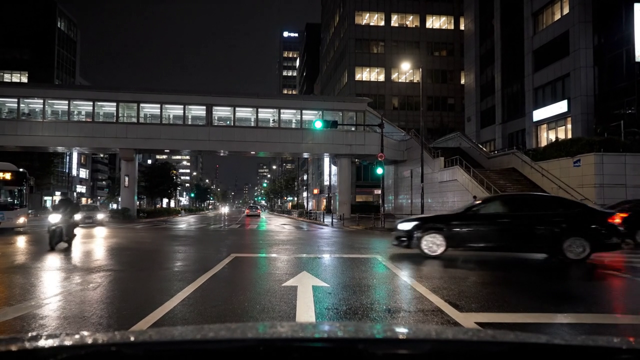} \\
                  & \footnotesize (a) structured image guided & \footnotesize (b) unstructured image guided & 
         \footnotesize (c) structured text guided & 
         \footnotesize (d) unstructured text guided  \\
    \end{tabular}
    \caption{Examples of video generation using the Veo3-preview WFM. For 3 different scenes of nuScenes-mini, we generate an 8 seconds video using the structured 5LM description as the prompt (structured) or a description generated without using the 5LM (unstructured). Each image correspond to the same frame, 4 seconds into the video. Those on the left were given the first frame of the original scene as a reference.}
    \label{fig:gen_table}

\end{figure*}

Using the generated scenario shown in Fig.~\ref{fig:L4_example}, we use Veo3~\cite{veo3} to generate video data (Fig.~\ref{fig:scene 1}). The prompt is our scenario description in the structured 5LM, only adding "\textit{From a dashcam point of view}" at the beginning. Fig.~\ref{fig:scene 1} show the generated scenes with (Fig.~\ref{subfig:L4_guided}) and without (Fig.~\ref{subfig:L4_non_guided}) using the reference image. Although the guided version shows better fidelity to the original, it also struggles with incoherence between the edited scenario and the reference image. Since the truck should be positioned in front of the ego vehicle but is not present on the first frame, it appears on the left of ego vehicle overtaking it. Fig.~\ref{fig:gen_table} shows the reconstruction with Veo 3 of three different scenes from nuScenes. When the generation is image-guided, the video generated using the 5LM shows better motion from the ego vehicle and other agents. When the video are not image-guided, while the background is more accurate to the original in the 5LM, there are more hallucinations (the pedestrian coming from the front instead of the side in S3). It is not clear at the moment if this is an issue with the structured 5LM, or with how the WFM should be prompted to respect large prompt.




%% file: sec/6_conclusion.tex
\section{Conclusion}
\label{sec:conclusion}

In this paper, we proposed a new synthetic scenario generation strategy focused on semantic descriptions to produce diverse road events. We showed how a structured representation format for driving scenarios can be used to create new editing strategies, but also for more detailed evaluation methods. Circumventing the application-subjective challenge of defining criteria for Edge Case identification, our method focuses on expending the operation domain of any given set of data. The evaluation metrics we propose are based on having a method to calculate the distance between any two scenarios. As such, they can be used with other types of structure as long as that condition is fulfilled. Although our focus is on rare driving scenarios, this allows our proposed method to be very flexible to many specific applications where such structure can be defined. Future works may explore further improvement of our method by exploring new structured models, especially in order to cover the temporal details of the scenes more precisely. Future research should pay attention to studying the behavior of LLM in regards to hallucination or catastrophic forgetting. Finally, the generation of visual data from text description should be improved to better ensure that the most relevant details of the scenes are preserved.


%% file: main.bbl
\begin{thebibliography}{10}
\providecommand{\url}[1]{#1}
\csname url@samestyle\endcsname
\providecommand{\newblock}{\relax}
\providecommand{\bibinfo}[2]{#2}
\providecommand{\BIBentrySTDinterwordspacing}{\spaceskip=0pt\relax}
\providecommand{\BIBentryALTinterwordstretchfactor}{4}
\providecommand{\BIBentryALTinterwordspacing}{\spaceskip=\fontdimen2\font plus
\BIBentryALTinterwordstretchfactor\fontdimen3\font minus \fontdimen4\font\relax}
\providecommand{\BIBforeignlanguage}[2]{{%
\expandafter\ifx\csname l@#1\endcsname\relax
\typeout{** WARNING: IEEEtran.bst: No hyphenation pattern has been}%
\typeout{** loaded for the language `#1'. Using the pattern for}%
\typeout{** the default language instead.}%
\else
\language=\csname l@#1\endcsname
\fi
#2}}
\providecommand{\BIBdecl}{\relax}
\BIBdecl

\bibitem{elhafsi2023semantic}
A.~Elhafsi, R.~Sinha, C.~Agia, E.~Schmerling, I.~A. Nesnas, and M.~Pavone, ``Semantic anomaly detection with large language models,'' \emph{Autonomous Robots}, vol.~47, no.~8, pp. 1035--1055, 2023.

\bibitem{dosovitskiy2017carla}
A.~Dosovitskiy, G.~Ros, F.~Codevilla, A.~Lopez, and V.~Koltun, ``Carla: An open urban driving simulator,'' in \emph{Conference on robot learning}.\hskip 1em plus 0.5em minus 0.4em\relax PMLR, 2017, pp. 1--16.

\bibitem{gao2025foundation}
Y.~Gao, M.~Piccinini, Y.~Zhang, D.~Wang, K.~Moller, R.~Brusnicki, B.~Zarrouki, A.~Gambi, J.~F. Totz, K.~Storms \emph{et~al.}, ``Foundation models in autonomous driving: A survey on scenario generation and scenario analysis,'' \emph{arXiv preprint arXiv:2506.11526}, 2025.

\bibitem{chang2024llmscenario}
C.~Chang, S.~Wang, J.~Zhang, J.~Ge, and L.~Li, ``Llmscenario: Large language model driven scenario generation,'' \emph{IEEE Transactions on Systems, Man, and Cybernetics: Systems}, vol.~54, no.~11, pp. 6581--6594, 2024.

\bibitem{10252672}
L.~Vater, M.~Sonntag, J.~Hiller, P.~Schaudt, and L.~Eckstein, ``A systematic approach towards the definition of the terms edge case and corner case for automated driving,'' in \emph{2023 3rd International Conference on Electrical, Computer, Communications and Mechatronics Engineering (ICECCME)}, 2023, pp. 1--6.

\bibitem{wang2024drivedreamer}
X.~Wang, Z.~Zhu, G.~Huang, X.~Chen, J.~Zhu, and J.~Lu, ``Drivedreamer: Towards real-world-drive world models for autonomous driving,'' in \emph{European Conference on Computer Vision}.\hskip 1em plus 0.5em minus 0.4em\relax Springer, 2024, pp. 55--72.

\bibitem{veo3}
{Google DeepMind}, ``Veo 3 model card,'' \url{https://storage.googleapis.com/deepmind-media/Model-Cards/Veo-3-Model-Card.pdf}, 2025.

\bibitem{aasi2024generating}
E.~Aasi, P.~Nguyen, S.~Sreeram, G.~Rosman, S.~Karaman, and D.~Rus, ``Generating out-of-distribution scenarios using language models,'' \emph{arXiv preprint arXiv:2411.16554}, 2024.

\bibitem{scholtes20216}
M.~Scholtes, L.~Westhofen, L.~R. Turner, K.~Lotto, M.~Schuldes, H.~Weber, N.~Wagener, C.~Neurohr, M.~H. Bollmann, F.~K{\"o}rtke \emph{et~al.}, ``6-layer model for a structured description and categorization of urban traffic and environment,'' \emph{iEEE Access}, vol.~9, pp. 59\,131--59\,147, 2021.

\bibitem{hu2022make}
Y.~Hu, C.~Luo, and Z.~Chen, ``Make it move: controllable image-to-video generation with text descriptions,'' in \emph{Proceedings of the IEEE/CVF Conference on Computer Vision and Pattern Recognition}, 2022, pp. 18\,219--18\,228.

\bibitem{ho2022video}
J.~Ho, T.~Salimans, A.~Gritsenko, W.~Chan, M.~Norouzi, and D.~J. Fleet, ``Video diffusion models,'' \emph{Advances in Neural Information Processing Systems}, vol.~35, pp. 8633--8646, 2022.

\bibitem{zhou2024storydiffusion}
Y.~Zhou, D.~Zhou, M.-M. Cheng, J.~Feng, and Q.~Hou, ``Storydiffusion: Consistent self-attention for long-range image and video generation,'' \emph{Advances in Neural Information Processing Systems}, vol.~37, pp. 110\,315--110\,340, 2024.

\bibitem{ni2023conditional}
H.~Ni, C.~Shi, K.~Li, S.~X. Huang, and M.~R. Min, ``Conditional image-to-video generation with latent flow diffusion models,'' in \emph{Proceedings of the IEEE/CVF conference on computer vision and pattern recognition}, 2023, pp. 18\,444--18\,455.

\bibitem{blattmann2023stable}
A.~Blattmann, T.~Dockhorn, S.~Kulal, D.~Mendelevitch, M.~Kilian, D.~Lorenz, Y.~Levi, Z.~English, V.~Voleti, A.~Letts \emph{et~al.}, ``Stable video diffusion: Scaling latent video diffusion models to large datasets,'' \emph{arXiv preprint arXiv:2311.15127}, 2023.

\bibitem{wu2024fairy}
B.~Wu, C.-Y. Chuang, X.~Wang, Y.~Jia, K.~Krishnakumar, T.~Xiao, F.~Liang, L.~Yu, and P.~Vajda, ``Fairy: Fast parallelized instruction-guided video-to-video synthesis,'' in \emph{Proceedings of the IEEE/CVF Conference on Computer Vision and Pattern Recognition}, 2024, pp. 8261--8270.

\bibitem{liang2024flowvid}
F.~Liang, B.~Wu, J.~Wang, L.~Yu, K.~Li, Y.~Zhao, I.~Misra, J.-B. Huang, P.~Zhang, P.~Vajda \emph{et~al.}, ``Flowvid: Taming imperfect optical flows for consistent video-to-video synthesis,'' in \emph{Proceedings of the IEEE/CVF Conference on Computer Vision and Pattern Recognition}, 2024, pp. 8207--8216.

\bibitem{fu2025llm}
Y.~Fu, R.~Zha, P.~Tian, and X.~Di, ``Llm-based realistic safety-critical driving video generation,'' \emph{arXiv preprint arXiv:2507.01264}, 2025.

\bibitem{gao2024vista}
S.~Gao, J.~Yang, L.~Chen, K.~Chitta, Y.~Qiu, A.~Geiger, J.~Zhang, and H.~Li, ``Vista: A generalizable driving world model with high fidelity and versatile controllability,'' \emph{Advances in Neural Information Processing Systems}, vol.~37, pp. 91\,560--91\,596, 2024.

\bibitem{hassan2024gem}
M.~Hassan, S.~Stapf, A.~Rahimi, P.~Rezende, Y.~Haghighi, D.~Br{\"u}ggemann, I.~Katircioglu, L.~Zhang, X.~Chen, S.~Saha \emph{et~al.}, ``Gem: A generalizable ego-vision multimodal world model for fine-grained ego-motion, object dynamics, and scene composition control,'' \emph{arXiv preprint arXiv:2412.11198}, 2024.

\bibitem{hu2023gaia1generativeworldmodel}
A.~Hu, L.~Russell, H.~Yeo, Z.~Murez, G.~Fedoseev, A.~Kendall, J.~Shotton, and G.~Corrado, ``Gaia-1: a generative world model for autonomous driving (2023),'' \emph{arXiv preprint arXiv:2309.17080}.

\bibitem{chang2025seeing}
C.-P. Chang, C.-Y. Wang, J.~Schmidt, H.~Caesar, and A.~Pagani, ``Seeing clearly, forgetting deeply: Revisiting fine-tuned video generators for driving simulation,'' \emph{arXiv preprint arXiv:2508.16512}, 2025.

\bibitem{caesar2020nuscenes}
H.~Caesar, V.~Bankiti, A.~H. Lang, S.~Vora, V.~E. Liong, Q.~Xu, A.~Krishnan, Y.~Pan, G.~Baldan, and O.~Beijbom, ``nuscenes: A multimodal dataset for autonomous driving,'' in \emph{Proceedings of the IEEE/CVF conference on computer vision and pattern recognition}, 2020, pp. 11\,621--11\,631.

\bibitem{russell2025gaia}
L.~Russell, A.~Hu, L.~Bertoni, G.~Fedoseev, J.~Shotton, E.~Arani, and G.~Corrado, ``Gaia-2: A controllable multi-view generative world model for autonomous driving,'' \emph{arXiv preprint arXiv:2503.20523}, 2025.

\bibitem{heusel2017gans}
M.~Heusel, H.~Ramsauer, T.~Unterthiner, B.~Nessler, and S.~Hochreiter, ``Gans trained by a two time-scale update rule converge to a local nash equilibrium,'' \emph{Advances in neural information processing systems}, vol.~30, 2017.

\bibitem{unterthiner2018towards}
T.~Unterthiner, S.~Van~Steenkiste, K.~Kurach, R.~Marinier, M.~Michalski, and S.~Gelly, ``Towards accurate generative models of video: A new metric \& challenges,'' \emph{arXiv preprint arXiv:1812.01717}, 2018.

\bibitem{blattmann2023align}
A.~Blattmann, R.~Rombach, H.~Ling, T.~Dockhorn, S.~W. Kim, S.~Fidler, and K.~Kreis, ``Align your latents: High-resolution video synthesis with latent diffusion models,'' in \emph{Proceedings of the IEEE/CVF conference on computer vision and pattern recognition}, 2023, pp. 22\,563--22\,575.

\bibitem{parisi2019continual}
G.~I. Parisi, R.~Kemker, J.~L. Part, C.~Kanan, and S.~Wermter, ``Continual lifelong learning with neural networks: A review,'' \emph{Neural networks}, vol. 113, pp. 54--71, 2019.

\bibitem{comanici2025gemini}
G.~Comanici, E.~Bieber, M.~Schaekermann, I.~Pasupat, N.~Sachdeva, I.~Dhillon, M.~Blistein, O.~Ram, D.~Zhang, E.~Rosen \emph{et~al.}, ``Gemini 2.5: Pushing the frontier with advanced reasoning, multimodality, long context, and next generation agentic capabilities,'' \emph{arXiv preprint arXiv:2507.06261}, 2025.

\bibitem{lee2025gemini}
J.~Lee, F.~Chen, S.~Dua, D.~Cer, M.~Shanbhogue, I.~Naim, G.~H. {\'A}brego, Z.~Li, K.~Chen, H.~S. Vera \emph{et~al.}, ``Gemini embedding: Generalizable embeddings from gemini,'' \emph{arXiv preprint arXiv:2503.07891}, 2025.

\end{thebibliography}
